\pdfoutput=1

\documentclass[11pt]{article}

\usepackage{acl}

\usepackage{times}

\usepackage{latexsym}
\usepackage[T1]{fontenc}

\usepackage[utf8]{inputenc}

\usepackage{microtype}

\usepackage{subcaption}

\usepackage{graphicx}
\usepackage{amsbsy}

\usepackage{amsmath}
\usepackage[ruled,vlined]{algorithm2e}
\SetInd{0.3em}{0.5em}
\SetArgSty{textup}

\usepackage{placeins}  
\usepackage{booktabs}
\usepackage[normalem]{ulem}
\useunder{\uline}{\ul}{}

\title{Using Pre-Trained Language Models for Producing Counter Narratives Against Hate Speech: a Comparative Study}

\author{Serra Sinem Tekiro\u{g}lu$^{2}$, Helena Bonaldi$^{1,2}$, Margherita Fanton$^{1,2}$\thanks{\indent Now at the University of Stuttgart, Germany.} , Marco Guerini$^{2}$  \\
  $^1$University of Trento, Italy \\
 $^2$Fondazione Bruno Kessler, Via Sommarive 18, Povo, Trento, Italy \\
  \texttt{tekiroglu@fbk.eu}, \texttt{hbonaldi@fbk.eu}, \\ \texttt{margherita.fanton@ims.uni-stuttgart.de},  \texttt{guerini@fbk.eu}}

\date{}

\begin{document}

\maketitle

\begin{abstract}
In this work, we present an extensive study on the use of pre-trained language models for the task of automatic Counter Narrative (CN) generation to fight online hate speech in English. We first present a comparative study to determine whether there is a particular Language Model (or class of LMs) and a particular decoding mechanism that are the most appropriate to generate CNs. Findings show that autoregressive models combined with stochastic decodings are the most promising. We then investigate how an LM performs in generating a CN with regard to an unseen target of hate.
We find out that a key element for successful `out of target' experiments is not an overall similarity with the training data but the presence of a specific subset of training data, i.\ e.\ a target that shares some commonalities with the test target that can be defined \textit{a-priori}. We finally introduce the idea of a pipeline based on the addition of an automatic post-editing step to refine generated CNs. 

\end{abstract}

\section{Introduction}
\noindent Hate Speech (HS) has found fertile ground in Social Media Platforms. Actions undertaken by such platforms to tackle online hatred consist in identifying possible sources of hate and removing them by means of content deletion, account suspension or shadow-banning. However, these actions are often interpreted and denounced as censorship by the affected users and 
political groups \citep{myers2018censored}. For this reason, such restrictions can have the opposite effect of exacerbating the hostility of the haters \citep{munger2017tweetment}. 
An alternative strategy, that is looming on the horizon, is based on the use of Counter Narratives. CNs are ``all communicative actions aimed at refuting hate speech through thoughtful and cogent reasons, and true and fact-bound arguments" \citep{schieb2016governing}. As a de-escalating measure, CNs have been proven to be successful in diminishing hate, while preserving the freedom of speech \citep{benesch2014countering, gagliardone2015countering}. An example of ${<}HS,CN{>}$ pair is shown below:\\

\noindent\textbf{HS}: Women are basically childlike, they remain this way most of their lives. Soft and emotional. It has devastated our once great patriarchal civilizations.\\
\noindent\textbf{CN}: \textit{Without softness and emotions there would be just brutality and cruelty. Not all women are soft and emotional and many men have these characteristics. To perpetuate these socially constructed gender profiles maintains norms which oppress anybody.}\\

\noindent Based on their effectiveness, CNs have started being employed by NGOs to counter online hate. Since for NGO operators it is impossible to manually write responses to all instances of hate, a line of NLP research has recently emerged, focusing on designing systems to automatically generate CN suggestions \citep{qian-etal-2019-benchmark, tekiroglu-etal-2020-generating, bonaldi2021data, chung-etal-2021-towards, zhu2021generate}.
In this study, our main goal is to compare pre-trained language models (LM) and decoding mechanisms in order to understand their pros and cons in generating CNs. Thus, we  use various automatic metrics and manual evaluations with expert judgments to assess several LMs, representing the main categories of the model architectures, and decoding methods. We further test the robustness of the fine-tuned LMs in generating CNs for an unseen target. Results show that autoregressive models are in general more suited for the task, and while stochastic decoding mechanisms can generate more novel, diverse, and informative outputs, the deterministic decoding is useful in scenarios where more generic and less novel (yet `safer') CNs are needed. Furthermore, in out-of-target experiments we find that the similarity of targets (e.g. \texttt{JEWS} and \texttt{MUSLIMS} as religious groups) plays a crucial role for the effectiveness of portability to new targets. We finally show a promising research direction of leveraging gold human edits for building an additional automatic post-editing step to correct errors made by LMs during generation. To the best of our knowledge, this is the first study systematically analysing state of the art pre-trained LMs in CN generation. 

\section{Related Work}\label{RelatedWork}
\noindent In this section we first discuss standard approaches to hate countering and studies on CN effectiveness on Social Media Platforms, then the existing CN data collection and generation strategies.

\paragraph{Hate countering.} NLP has started addressing the phenomenon of the proliferation of HS by creating datasets  for automatic detection \citep{mathew2021hatexplain, cao2020deephate, kumar2018benchmarking,hosseinmardi2015detection, waseem2016you, burnap2016us}. Several surveys provide a review on the existing approaches on the topic \citep{poletto2020resources, schmidt2017survey, nunes2018survey}, also addressing the ethical challenges of the task \citep{kiritchenko2020confronting}.
Still, automatic detection of HS presents some drawbacks \citep{vidgen2020directions}. First of all, the datasets might include biases, and the models tend to replicate such biases \citep{10.1007/978-3-319-67256-4_32, davidson2019racial, sap2019risk, tsvetkov2020demoting}. 
Moreover, the end goals for which HS detection is employed are often charged with censorship of the freedom of speech by concerned users \citep{munger2017tweetment, myers2018censored}. In this scenario, NGOs have started employing CNs to counter online hate. 
CNs have been shown to be effective in reducing linguistic violence \citep{benesch2014countering, gagliardone2015countering, schieb2016governing, silverman2016impact, mathew2019thou}; moreover, even if they might not influence the view of extremists, they are still effective in presenting alternative and non-hateful viewpoints to bystanders \citep{allison2016cyber,anderson2014combating}. 

\paragraph{CN data collection.} 
The existing studies for collecting CN datasets employ four main approaches. \textit{Crawling} consists in automatically scraping websites, starting from an HS content and searching for possible CNs among the responses \citep{mathew2018analyzing, mathew2019thou}. With \textit{crowdsourcing} CNs are written by non-expert paid workers as responses to provided hate content \citep{qian-etal-2019-benchmark}. \textit{Nichesourcing} relies on a niche group of experts for data collection \citep{de2012nichesourcing}, and it was employed by \citet{conan-2019} for CN collection using NGO's operators. \textit{Hybrid approaches} use a combination of LMs and humans to collect data \citep{GAN_hum_loop, dinan2019build, vidgen2020learning}. Studies on CN collection are presented in more detail by \citet{tekiroglu-etal-2020-generating, bonaldi2021data}.

\paragraph{CN generation.}  
Neural approaches to automatically generate CNs are beginning to be investigated. \citet{bonaldi2021data, tekiroglu-etal-2020-generating, qian-etal-2019-benchmark} employ a mix of automatic and human intervention to generate CNs. \citet{zhu2021generate} propose an entirely automated pipeline of candidate CN generation and filtering. Other lines of work include CN generation for under-resourced languages such as for Italian \citep{chung2020italian}, and the generation of knowledge-bound CNs, which allows the production of CNs based on grounded and up-to-date facts and plausible arguments, avoiding the hallucination phenomena \citep{chung-etal-2021-towards}. Instead, in our work we take a more foundational perspective, which is relevant for all the LM-based pipelines described above. Therefore, we compare and assess various state of the art pre-trained LMs in an end-to-end setting, which is developed as a downstream task for CN generation.

\section{Methodology}\label{ch:meth}

In this section, we present the CN dataset, the language models, and the decoding mechanisms employed for our experiments. 

\subsection{Dataset for fine-tuning}\label{sec:dataset}

For this study we rely on the dataset proposed by \citet{bonaldi2021data}, 
which is the only available dataset that grants both the target diversity and the CN quality we aim for. The dataset was collected with a human-in-the-loop approach, by employing an autoregressive LM (GPT-2) paired with three expert human reviewers. It features 5003 ${<}HS,CN{>}$ pairs, covering several targets of hate including \texttt{DISABLED}, \texttt{JEWS}, \texttt{LGBT+}, \texttt{MIGRANTS}, \texttt{MUSLIMS}, \texttt{POC}, \texttt{WOMEN}. The residual categories are collapsed to the label \texttt{OTHER}. We partitioned the dataset into training, validation, and test sets with the ratio: $8:1:1$ (i.\ e.\ 4003, 500 and 500 pairs), ensuring that all sets share the same target distribution, and no repetition of HS across the sets is allowed.

\subsection{Models} \label{ch:meth-models}

\noindent We experiment with 5 Transformer based LMs \citep{vaswani2017attention} representing the main categories of the model mechanisms: autoregressive, autoencoder, and seq2seq.

\noindent\textbf{BERT.}
The Bidirectional Encoder Representations from Transformers was introduced by \citet{devlin2019bert}. It is a bidirectional autoencoder that can be adapted to text generation \citep{wang2019bert}. 

\noindent\textbf{GPT-2.}
The Generative Pre-trained Transformer 2 is an autoregressive model built for text generation \citep{radford2019language}. 

\noindent\textbf{DialoGPT.}
The Dialogue Generative Pretrained Transformer is the extension of GPT-2 specifically created for conversational response generation \citep{zhang2020dialogpt}. 

\noindent\textbf{BART.}
BART is a denoising autoencoder for pre-training seq2seq models \citep{lewis2020bart}. The encoder-decoder architecture of BART is composed of a bidirectional encoder and an autoregressive decoder. 

\noindent\textbf{T5.} The Text-to-Text Transfer Transformer proposed by \citet{raffel2020t5} is a seq2seq model with an encoder-decoder Transformer architecture.\\

While all the other models could be fine-tuned directly for the generation task, for BERT we warmstarted an encoder-decoder model using BERT checkpoints similar to the BERT2BERT model defined by \citep{rothe2020leveraging}. The fine-tuning details and hyperparameter settings can be found in Appendix \ref{sec:AppendixA}.

\subsection{Decoding mechanisms}\label{sec:decodings}

We utilize 4 decoding mechanisms: a deterministic (Beam Search) and three stochastic (Top-$k$, Top-$p$,  and a combination of the two).

\noindent\textbf{Beam Search (BS).} The Beam Search algorithm is designed to pick the most-likely sequence 
\citep{li-etal-2016-deep, wiseman-etal-2017-challenges}.

\noindent\textbf{Top-$\boldsymbol{k}$ (\textbf{Top$\boldsymbol{_k}$}).} The sampling procedure proposed by \citet{fan-etal-2018-hierarchical} selects a random word from the $k$ most probable ones, at each time step. 

\noindent\textbf{Top-$\boldsymbol{p}$ (\textbf{Top$\boldsymbol{_p}$}).} Also known as Nucleus Sampling, the parameter $p$ indicates the total probability for the pooled candidates, at each time step \citep{holtzman2019curious}. 

\noindent\textbf{Combining Top-$\boldsymbol{p}$ and Top-$\boldsymbol{k}$ (\textbf{Top$\boldsymbol{_{pk}}$}).}
At decoding stage, it is possible to combine the parameters $p$ and $k$. This is a Nucleus Sampling constrained to the Top-$k$ most probable words.

In our experiments we used the following parameters as default \citep{wiseman-etal-2017-challenges,holtzman2019curious}: Beam-Search with 5 beams and repetition penalty = 2; Top-$k$ with $k = 40$; Top-$p$ with $p = .92$; Top$_{pk}$ with $k = 40$ and $p = .92$.

\section{Evaluation metrics}\label{sec:EvaluationMetrics}

We use several metrics to evaluate various aspects of the CN generation.

\noindent{\textbf{Overlap Metrics.}} 
These metrics depend on the $n$-gram similarity of the generated outputs to a set of reference texts in order to assess the quality. We used our gold CNs as \textit{references} and the CNs generated by the different models, as \textit{candidates}. In particular, we employed three BLEU variants: BLEU-1 (B-1), BLEU-3 (B-3) and BLEU-4 (B-4) \citep{papineni2002bleu}, and ROUGE-L (ROU) \citep{lin2004rouge}.

\noindent{\textbf{Diversity metrics.}} They are used to measure how diverse and novel the produced CNs are. In particular, we utilized \textit{Repetition Rate} (RR) to measure the repetitiveness across generated CNs, in terms of the average ratios of non-singleton $n$-grams present in the corpus \citep{bertoldi2013cache}. It should be noted that RR is calculated as a corpus-based repetition score , i.e. inter-CN, instead of calculating intra-CN repetition of $n$-grams only.

We also used \textit{Novelty} (NOV) \citep{wang2018sentigan}, based on Jaccard similarity, to compute the amount of novel content that is present in the generated CNs as compared to the training data. 

\noindent{\textbf{Human evaluation metrics.}}\label{humaneval} Albeit more difficult to attain, human judgments provide a more reliable evaluation and a deeper understanding than automatic metrics \citep{belz2006comparing, novikova2017we}. To this end, we specified the following dimensions for the evaluation of CNs.
\textit{Suitableness} (SUI): measures how suitable a CN is to the HS in terms of semantic relatedness and in terms of adherence to CN guidelines\footnote{See for example \url{https://getthetrollsout.org/stoppinghate}}; \textit{Grammaticality} (GRM): how grammatically correct a generated CN is; \textit{Specificity} (SPE): how specific are the arguments brought by the CN in response to the HS; \textit{Choose-or-not} (CHO): determines whether the annotators would select that CN to post-edit and use it in a real case scenario as in the set up presented by  \citet{chung2021empowering}; \textit{Is-best} (BEST): whether the CN is the absolute best among the ones generated for an HS (i.\ e.\ whether the annotators would pick up exactly that CN if they had to use it in a real case scenario).

The first three dimensions are rated with a 5-points Likert scale and follow the evaluation procedure described by \citet{chung2020italian}, whereas both choose-or-not and is-best are binary ratings (0, 1). Choose-or-not allows for multiple CNs to be selected for the same HS, while only one CN can be selected for is-best for each HS.

\noindent{\textbf{Toxicity.}\footnote{\url{https://www.perspectiveapi.com}}} It determines how ``rude, disrespectful, or unreasonable'' a text is. Toxicity has been employed both to detect the bias present in LMs \citep{gehman2020realtoxicityprompts} and as a solution to mitigate such bias \citep{gehman2020realtoxicityprompts, xu2020recipes}.

\noindent\textbf{Syntactic metrics.} A high syntactic complexity can be used as a proxy for an LM's ability of generating complex arguments. We used the syntactic dependency parser of spaCy\footnote{\url{https://spacy.io/usage/linguistic-features\#dependency-parse}} For the task, focusing on the following measures: \textit{Maximum Syntactic Depth} (MSD): the maximum depth among the dependency trees calculated over each sentence composing a CN. \textit{Average Syntactic Depth} (ASD): the average depth of the sentences in each CN. \textit{Number of Sentences} (NST): the number of sentences composing a CN. 

\section{Experiments}\label{ch:exper}
We performed two sets of experiments: first, we assessed how LMs perform in the task of generating CNs with different decoding mechanisms. Then, we selected the best model from the first round of experiments and tested its generalization capabilities when confronted with an unseen target of hate.

\subsection{LMs and decoding experiments}\label{sec:exp1}

For the first round of experiments, in order to avoid possible unfair assessments given by the open nature of the generative task
(i.\ e.\ a highly suitable CN candidate could be scored low due to its difference from the single reference/gold CN), at test time we allowed the generation of several candidates for each HS+LM+decoding mechanism combination. We loosely drew inspiration from the Rank-$N$ Accuracy procedure and the `generate, prune, select' procedure \citep{zhu2021generate}. In particular, given an LM and a decoding mechanism, we generated 5 CNs for each HS in the test set.

\begin{center}
\begin{figure}[htbp!]\centering{\includegraphics[width=0.97\columnwidth]{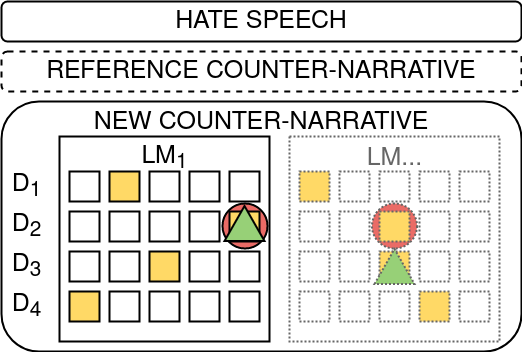}}

\caption{Given an HS, 5 CNs are generated for each model-decoding combination. \includegraphics[scale=0.2]{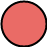} indicates the best CN per model ($\in$ Best$_\text{LM}$). \includegraphics[scale=0.2]{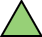} indicates the best CN per decoding ($\in$ Best$_\text{D}$).  \includegraphics[scale=0.25]{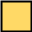} indicates the best CN per model-decoding combination ($\in$ Best$_\text{LM+D}$).}
\label{fig:Experiment1_new2}
\end{figure}\end{center}

\paragraph{Automated evaluation and selection}
We set up the automatic evaluation strategy as displayed in Figure \ref{fig:Experiment1_new2}. First, we scored each CN with the overlap metrics presented in Section \ref{sec:EvaluationMetrics}, using the gold CN as a reference. Next, we ranked the candidate CNs with respect to the overlap scores and computed the mean of the rankings. Then, we selected the \textit{best} ones according to the following criteria:

\noindent \textbf{Best$_\text{LM}$} 
selects the single best CN for an HS among the 20 generated by the 4 models. 

\noindent \textbf{Best$_\text{D}$} 
selects the single best CN for an HS among the 25 generated by the 5 decoding configurations.

\noindent \textbf{Best$_\text{LM+D}$} 
selects the single best CN among the 5 generated with each model-decoding combination. 

\noindent Moreover, we assessed the overall corpus-wise quality of the generated CNs with respect to the models, to the decoding mechanisms, and to the model-decoding combinations via the diversity metrics.

\paragraph{Human evaluation on a sample}
To perform the human evaluation we referred to the Best$_{\text{\textsc{LM}}}$ generations and sampled 200 instances from it. Each instance comprises an HS and 5 relevant CNs, each generated by a different model. 
We recruited 2 annotators who were trained extensively for the task following the procedure used by \citet{bonaldi2021data}.
The expert annotators were asked to evaluate the 5 CNs corresponding to the HS, according to the dimensions described in Section \ref{humaneval}.  
We enriched the evaluation of this subset with the toxicity and the syntactic metrics.

\subsection{Results of the first set of experiments} \label{ch:results-experiment1}
The results of the experiments on the LMs and the decoding mechanisms are reported in this section\footnote{The training details for all the models we employed are described in Appendix \ref{sec:AppendixA}}.

\paragraph{Best Model}
The results of the comparison of the models on the Best$_{\text{\textsc{LM}}}$ generations can be found in Table \ref{tab:ex1_models}. Regarding the overlap and diversity metrics, DialoGPT records
the best or the second best score in all the metrics, apart from novelty where it still achieves a high score (0.643) close to the best performance (0.655). T5 also achieves high scores, especially on ROUGE, BLEU-1 and novelty.
\begin{table*}[htbp]
\begin{center}\resizebox{\textwidth}{!}{
\begin{tabular}{l|rrrr|rr|r|rrr|rrrrr}
\hline
\textbf{} & \multicolumn{4}{c}{\textbf{Overlap}}                                                                                                         & \multicolumn{2}{|c}{\textbf{Diversity}}                         & \multicolumn{1}{|c}{\textbf{Toxicity}} & \multicolumn{3}{|c}{\textbf{Syntactic metrics}}                                                         & \multicolumn{5}{|c}{\textbf{Human evaluation}}                                                                         \\ \hline
          & \multicolumn{1}{c}{\textbf{ROU}} & \multicolumn{1}{c}{\textbf{B-1}} & \multicolumn{1}{c}{\textbf{B-3}} & \multicolumn{1}{c}{\textbf{B-4}} & \multicolumn{1}{|c}{\textbf{RR}} & \multicolumn{1}{c}{\textbf{NOV}} & \multicolumn{1}{|c}{\textbf{-}}        & \multicolumn{1}{|c}{\textbf{ASD}} & \multicolumn{1}{c}{\textbf{MSD}} & \multicolumn{1}{c|}{\textbf{NST}} & \textbf{SUI}         & \textbf{SPE}         & \textbf{GRM}        & \textbf{CHO}       & \textbf{BEST}         \\
BART      & 0.268                             & 0.277                              & 0.085                              & \textbf{0.051}                     & 20.722                         & 0.560                               & 0.420                                & 4.311                           & 4.965                           & 1.740                           & {\ul \textbf{3.790}} & 2.552                & {\ul \textbf{4.937}} & {\ul \textbf{0.840}} & {\ul \textbf{0.272}} \\
BERT      & 0.237                             & 0.277                              & 0.073                              & 0.037                              & 24.747                         & 0.605                               & 0.406                                & {\ul \textbf{5.008}}            & {\ul \textbf{6.160}}            & {\ul \textbf{2.280}}            & 3.135                & 2.647                & 4.247                & 0.717                & 0.122                \\
T5        & {\ul \textbf{0.274}}              & \textbf{0.302}                     & 0.083                              & 0.042                              & 8.548                          & {\ul \textbf{0.655}}                & 0.359                                & \textbf{4.692}                  & 5.325                           & 1.715                           & 2.872                & 2.402                & 4.680                & 0.642                & 0.090                \\
DialoGPT  & \textbf{0.273}                    & {\ul \textbf{0.304}}               & {\ul \textbf{0.093}}               & {\ul \textbf{0.052}}               & \textbf{8.248}                 & 0.643                               & \textbf{0.343}                       & 4.677                           & 5.575                           & 1.895                           & 3.392                & \textbf{2.755}       & \textbf{4.880}       & 0.767                & 0.245                \\
GPT-2     & 0.264                             & 0.297                              & \textbf{0.088}                     & 0.050                              & {\ul \textbf{7.736}}           & \textbf{0.653}                      & {\ul \textbf{0.342}}                 & 4.584                           & \textbf{5.595}                  & \textbf{2.240}                  & \textbf{3.555}       & {\ul \textbf{2.880}} & 4.867                & \textbf{0.795}       & \textbf{0.270}       
\end{tabular}}
\caption{Results of the overlap and diversity metrics are calculated on the Best$_{\text{\textsc{LM}}}$ generations while the toxicity, the syntactic metrics and the human evaluation are calculated on the corresponding subset.
}
\label{tab:ex1_models}
\end{center}
\end{table*}

BART, instead, is the best model according to human evaluation metrics, apart from specificity.
On the other hand, it shows poor performances in terms of diversity metrics, indicating that it tends to produce grammatical and suitable but very generic responses. 

BERT records the worst scores for all the overlap and diversity metrics apart from novelty. However, it also achieves the best syntactic metric results. Therefore, it is evident that BERT's output is more complex, but very repetitive. The combination of these aspects eventually affects the clarity of BERT's output such that it yields poor results in the human evaluation, in particular for grammaticality (4.2, while other models are above 4.6).  This poor grammaticality can also explain the syntactic scores since the spaCy dependency parser was not trained to handle ungrammatical text and this could actually inflates the ASD and MSD scores.

GPT-2 overall yields very competitive results for several groups of metrics. It obtains the second-highest novelty score (0.653) and the best RR (7.736). It also achieves the second best results on BLEU-3, maximum syntactic depth and number of sentences, and the best results on toxicity and specificity (2.880) indicating the ability to produce complex, suitable, focused and diverse CNs.

After the human evaluation we ran a qualitative interview with the annotators, whose feedback on the data strengthened the results we observed and the conclusion we drew. For instance, they reported the repetition of simple, yet catch-them-all, expressions (e.g. ``they are our brothers and sisters") regardless of the target. Further inspections found that those CNs were mainly produced by BERT, which is in line with BERT's RR score. 

\paragraph{Best Decoding mechanism.}
The results calculated on Best$_{\text{\textsc{D}}}$ output are presented in Table \ref{tab:ex1_decodings}. Top$_k$ is the best performing decoding mechanism achieving the best results on the diversity metrics, BLEU-3 and BLEU-4. It is also the best performing for specificity, maximum syntactic depth and number of sentences, and the second best for average syntactic depth and toxicity.

\begin{table*}[htbp]
\begin{center}\resizebox{\textwidth}{!}{
\begin{tabular}{l|lrrr|rr|r|rrr|lrrrr|r}
\hline
\textbf{}  & \multicolumn{4}{|c}{\textbf{Overlap}}                                                                                                      & \multicolumn{2}{|c}{\textbf{Diversity}}                             & \multicolumn{1}{|c}{\textbf{Toxicity}} & \multicolumn{3}{|c}{\textbf{Syntactic metrics}}                                                         & \multicolumn{5}{|c}{\textbf{Human evaluation}} 
 & \multicolumn{1}{|c}{}     \\ \hline
           & \multicolumn{1}{|c}{\textbf{ROU}} & \multicolumn{1}{c}{\textbf{B-1}} & \multicolumn{1}{c}{\textbf{B-3}} & \multicolumn{1}{c}{\textbf{B-4}} & \multicolumn{1}{|c}{\textbf{RR}} & \multicolumn{1}{c}{\textbf{NOV}} & \multicolumn{1}{|c}{\textbf{-}}        & \multicolumn{1}{|c}{\textbf{ASD}} & \multicolumn{1}{c}{\textbf{MSD}} & \multicolumn{1}{c}{\textbf{NST}} & \multicolumn{1}{|c}{\textbf{SUI}} & \multicolumn{1}{c}{\textbf{SPE}} & \multicolumn{1}{c}{\textbf{GRM}} & \multicolumn{1}{c}{\textbf{CHO}} & \multicolumn{1}{c}{\textbf{BEST}} & \multicolumn{1}{|c}{\textbf{n}} 
           \\
BS         & {\ul \textbf{0.287}}             & 0.299                            & 0.096                            & 0.059                            & 21.579                          & 0.561                            & 0.398                                 & 4.415                            & 5.048                            & 1.684                            & {\ul \textbf{3.936}}              & 2.497                             & {\ul \textbf{4.925}}               & {\ul \textbf{0.826}}                & {\ul \textbf{0.222}}              & \%18.7                            \\
Top$_{pk}$ & \textbf{0.287}                   & {\ul \textbf{0.320}}             & {\ul \textbf{0.106}}             & 0.059                            & 11.404                          & 0.639                            & {\ul \textbf{0.352}}                  & 4.676                            & 5.488                            & 1.932                            & \textbf{3.324}                    & 2.647                             & \textbf{4.688}                     & \textbf{0.764}                      & \textbf{0.212}                    & \%29.3                            \\
Top$_k$    & 0.282                            & 0.314                            & {\ul \textbf{0.106}}             & {\ul \textbf{0.060}}             & {\ul \textbf{10.076}}           & {\ul \textbf{0.652}}             & \textbf{0.374}                        & \textbf{4.704}                   & {\ul \textbf{5.756}}             & {\ul \textbf{2.133}}             & 3.155                             & {\ul \textbf{2.716}}              & 4.659                              & 0.716                               & 0.183                             & \%27.1                            \\
Top$_p$    & 0.285                            & \textbf{0.319}                   & 0.105                            & {\ul \textbf{0.060}}             & \textbf{11.270}                 & \textbf{0.640}                   & 0.381                                 & {\ul \textbf{4.753}}             & \textbf{5.671}                   & \textbf{2.068}                   & 3.149                             & \textbf{2.687}                    & 4.681                              & 0.723                               & 0.189                             &\%24.9                   

\end{tabular}}
\end{center}
\caption{The results for the overlap and diversity metrics are calculated on the Best$_{\text{\textsc{D}}}$ generations: for each decoding mechanism, there are 2500 CNs. The remaining metrics are calculated on a subset of 1000 CNs: the distribution of which is shown in the column "n".}
\label{tab:ex1_decodings}
\end{table*}

The other stochastic decoding mechanisms perform well too.
Top$_p$ yields competitive results on both diversity and overlap metrics; it is the second best for specificity, and achieves good results on the syntactic metrics. Top$_{pk}$ has a good performance on the overlap metrics. It obtains the second-highest scores in most of the human evaluation metrics and the lowest in toxicity, and it reaches a reasonable specificity score.

On the other hand, BS does not achieve particularly good results, except for the ROUGE score. Even if it is the best decoding with respect to the
human evaluation, this comes at the cost of specificity and diversity. Through a post-hoc manual analysis we observed that it was due to the deterministic nature of BS, that tends to choose the most probable sequences, i.\ e.\ the ``safest", thus resulting in vague and repetitive outputs. 

\paragraph{Best Model-Decoding combination} Here we briefly discuss the results of the evaluation obtained on the Best$_{\text{\textsc{lm+d}}}$ generations. 
In particular, the autoregressive models GPT-2 and DialoGPT behave similarly with similar decoding mechanisms, such that BS outputs the best results for almost all the overlap metrics, and the worst for the diversity metrics. On the contrary, for the other models, the results achieved with stochastic decoding mechanisms are the best for the overlap metrics. In almost all the cases, we observe that the stochastic decoding mechanisms perform better on syntactic and diversity metrics and on toxicity, while for the human evaluation metrics BS tends to be the best, except for specificity. A detailed discussion can be found in Appendix \ref{sec:AppendixD}.

\paragraph{Discussion.}
In this set of experiments, we found that the autoregressive models perform the best according to a combination of several metrics that we deem particularly relevant (e.g. more novel, diverse, and informative outputs). Of course more repetitive and conservative outputs can be preferred when high precision of suitable CNs are required at the expense of being more generic and less novel.
Still, for what concerns autoregressive models it could be argued that the good performance of the GPT-2 model we fine-tuned is due to the fact that generated CNs and gold CNs derive from a similar distribution (GPT-2 was employed in the human-in-the-loop process used to create the reference dataset from  \citet{bonaldi2021data}). While we recognize that this could partially explain the performance of our GPT-2 model, it does not explain the performance of DialoGPT, which is pre-trained on a completely different dataset. Therefore, we can reasonably conclude that autoregressive models are particularly suited for the task, regardless of the pre-training data. 

With respect to the decoding mechanisms, we record high repetitiveness and low novelty for the deterministic decoding BS. Even if it reaches high scores in most of the human evaluation metrics, it fails to produce specific CNs ending up in generating suitable, yet generic responses. On the contrary, stochastic decoding mechanisms produce more novel and specific responses. 

Example CNs generated in this session of experiments, along with some qualitative analysis, can be found in Appendix \ref{sec:AppendixC}.

\subsection{Leave One Target Out experiments}\label{subsec:loto-experiment}
In the second stage, we built a set of cross-domain experiments to capture the generalization capabilities of the best LM determined in the previous experiments. Specifically, we concentrate on assessing how much a pre-trained language model fine-tuned on a pool of hate targets can generalize to an unseen target.

Thus, for the out of target experiment we selected the LM that we deem the most prominent in order to reduce the number of LM configurations to compare. In particular, since we want to examine the generalization capability of the LM, the generation of \textit{novel} CNs, in comparison to the training data, is given primary importance. Secondly, \textit{specificity} is also crucial since it signifies the ability of the LM/decoding mechanism in generating accurate CNs and avoiding vague yet suitable, catch-all CNs. In contrast, repetitiveness is an undesirable feature of CNs, as it signals the tendency of a model to produce less flexible content. 
Given these considerations, we chose to employ GPT-2 with Top$\boldsymbol{_k}$ decoding for the Leave One Target Out (\textsc{loto}) experiments  
since it is the configuration achieving the best trade-off amongst all the others. 

This set of experiments is structured in 3 steps, replicated for each of the selected targets. We selected the targets with the highest number of examples (\texttt{MUSLIMS}, \texttt{MIGRANTS}, \texttt{WOMEN}, \texttt{LGBT+} and \texttt{JEWS}) to have a sufficient sized test set for each configuration.  

First, we sampled from the \citet{bonaldi2021data} dataset 600 pairs for each \textsc{loto} target, in order to have a balanced setting. Additionally, \texttt{POC} and \texttt{DISABLED} were always kept in the training set, and we removed multi-target cases from \texttt{OTHER}. The resulting dataset consists of 3729 instances (further details are provided in Appendix \ref{sec:AppendixB}). 
Secondly, we fine-tuned 5 different configurations of the LM, and in each configuration one of the 5 \textsc{loto} targets is not present in the training data: 
LM$\boldsymbol{_{\text{-JEWS}}}$, LM$\boldsymbol{_{\text{-LGTB+}}}$, LM$\boldsymbol{_{\text{-MIGRANTS}}}$, LM$\boldsymbol{_{\text{-MUSLIMS}}}$ and LM$\boldsymbol{_{\text{-WOMEN}}}$.
Finally, we tested each \textsc{loto} model on the 600 HSs in the test set made of ``left out" target examples. For instance, the model LM$\boldsymbol{_{\text{-JEWS}}}$ is used for generating the CNs for the target \texttt{JEWS}, after being trained on ${<}HS,CN{>}$ data without any instances with the label \texttt{JEWS}. We generated 5 CNs for each HS and selected the best CN according to the procedure described in Section \ref{sec:exp1}.

\subsection*{Results of \textsc{loto} experiments} \label{subsec:loto-results}

We analyse the CNs generated with the \textsc{loto} models through overlap and diversity metrics (Table \ref{tab:LOTO_results}). We refer to Appendix \ref{sec:AppendixB} for the comparison between RR calculated on the candidate CNs and the reference CNs of the \citet{bonaldi2021data} dataset.

For all the targets we record higher novelty scores as compared to the previous experiments. Higher novelty ranges indicate that conditioning with new material (i.\ e.\ HS for the unseen targets) induces GPT-2 to produce new arguments.
On the other hand, as expected, the overlap scores for \textsc{loto} are remarkably lower than those from the previous experiments (Table \ref{tab:LOTO_results}). Therefore, we can infer that generalizing to an unseen target is harder than generalizing to an unseen HS.  

\begin{table}[htbp]
\begin{center}\resizebox{\columnwidth}{!}{
\begin{tabular}{@{}l|rrrr|rr@{}}  \hline
\textsc{loto}     & \multicolumn{4}{|c}{\textbf{Overlap}}                                                                                                    & \multicolumn{2}{|c}{\textbf{Diversity}}                                                             \\ \hline
Target  & \multicolumn{1}{l}{\textbf{ROU}} & \multicolumn{1}{l}{\textbf{B-1}} & \multicolumn{1}{l}{\textbf{B-3}} & \multicolumn{1}{l}{\textbf{B-4}} & \multicolumn{1}{|l}{\textbf{RR}} &  \multicolumn{1}{l}{\textbf{NOV}} \\
\texttt{JEWS}    & 0.1609                             & 0.1842                              & 0.0134                              & 0.0035                              & 4.796                                                   & 0.718                                \\
\texttt{LGBT+}   & 0.1599                             & 0.1828                              & 0.0149                              & 0.0055                              & {\ul \textbf{4.620}}                                    & 0.718                                \\
\texttt{MIGRANTS}& 0.1659                             & 0.1915                              & 0.0163                              & 0.0038                              & 4.707                                                 & \textbf{0.720}                       \\
\texttt{MUSLIMS}  & \textbf{0.1743}                    & \textbf{0.1934}                     & {\ul \textbf{0.0197}}               & \textbf{0.0059}                     & 5.314                                              & 0.712                                \\
\texttt{WOMEN}    & {\ul \textbf{0.1755}}              & {\ul \textbf{0.1988}}               & \textbf{0.0195}                     & {\ul \textbf{0.0068}}               & \textbf{4.632}                                         & {\ul \textbf{0.729}}                 
\end{tabular}}
\end{center}
\caption{The overlap and diversity metrics scores for the various \textsc{loto} configurations.}
\label{tab:LOTO_results}
\end{table}

We also found out that the CNs generated in the LM$\boldsymbol{_{\text{-MUSLIMS}}}$ and  LM$\boldsymbol{_{\text{-WOMEN}}}$ configurations obtain the highest overlap scores (Table \ref{tab:LOTO_results}). We hypothesize that the high scores can be explained by the presence of a target in the \textsc{loto} training that is highly similar to the left out one. To this end, we computed the novelty between each target subset of the training data and the \textsc{loto} test data for that configuration (see Appendix \ref{sec:AppendixB} for details).

The reference CNs for LM$\boldsymbol{_{\text{-MUSLIMS}}}$ record the lowest novelty scores with respect to the \texttt{JEWS} subset of the training set (i.\ e.\ 0.761). Thus, it can be interpreted as the most influential portion of training data for the target \texttt{MUSLIMS}. On the other hand, for LM$\boldsymbol{_{\text{-WOMEN}}}$ the highest influence is recorded with the \texttt{LGBT+} subset of the training data (i.\ e.\ 0.763). These results can be explained by the semantic similarity of the target \texttt{MUSLIMS} to \texttt{JEWS}, both being religious groups; and of \texttt{WOMEN} to \texttt{LGBT+}, both being related to gender issues. 

\begin{figure}[t!]
\centering
\begin{subfigure}{1\columnwidth}
{\includegraphics[width=1\columnwidth]{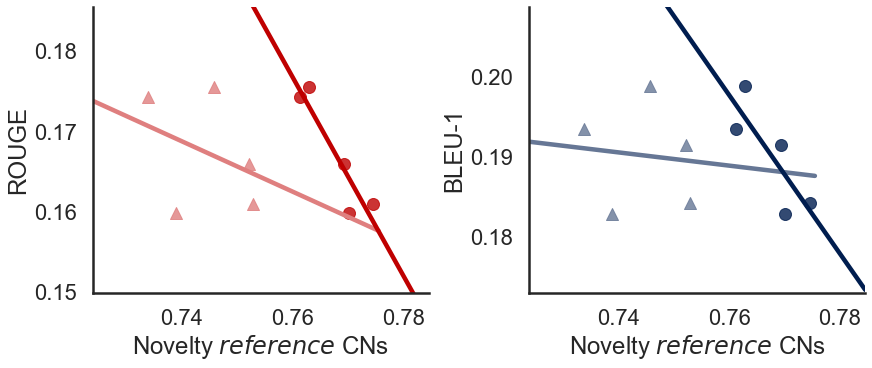}}

\end{subfigure}
\begin{subfigure}{1\columnwidth}
{\includegraphics[width=1\columnwidth]{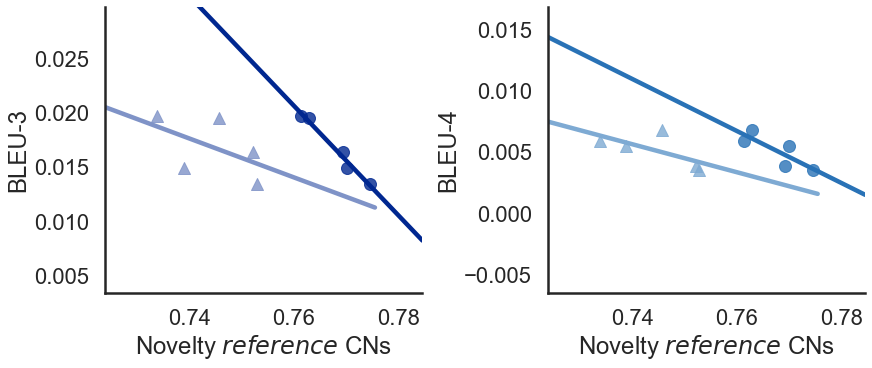}}

\end{subfigure}
\caption{The correlation between the novelty of the reference CNs and overlap metrics: in each plot, the dots and the darker line correspond to the most influential target; the triangles and the lighter line correspond to the results calculated without it.}
\label{fig:correlation_novelty_quality}
\end{figure}

As a complementary analysis, we consider two different computations of the reference CN novelty: with respect to the most influential target for each \textsc{loto} configuration, and with respect to the \textsc{loto} training data without the most influential target. We computed the Pearson correlation between the overlap metrics and each of the two novelty computations.
In Figure \ref{fig:correlation_novelty_quality}, we observe that removing the influential target from the training data strongly decreases the correlation with the overlap metrics (from an average of -0.889 to -0.416). Consequently, we can conclude that to obtain high overlap results in the \textsc{loto} experiments, it is necessary that the training data contains a target strongly connected to the left out one. Most importantly, this connection is not arbitrarily decided but it is based on an \textit{a-priori} semantic similarity of the targets as exemplified before. 

\begin{center}
\begin{figure}[htbp!]{\includegraphics[width=1\columnwidth]{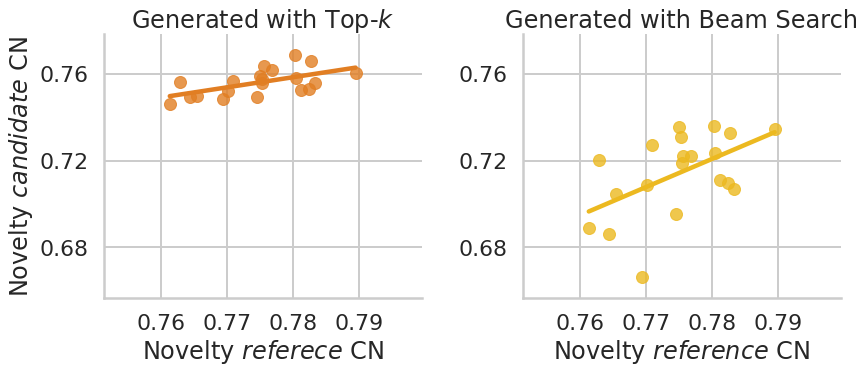}}
\caption{Reference and candidate CNs novelty, for Top-$k$ and BS \textsc{loto} generations.}
\label{fig:correlation_topk_bs}
\end{figure}\end{center}

Finally, we chose to generate also with the BS decoding mechanism, to use it as a baseline and compare it to the stochastic decoding mechanism (Top-$k$). In particular, we computed the Pearson correlation between the novelty of the reference CNs and the novelty of the candidate CNs with respect to the corresponding training data (Figure \ref{fig:correlation_topk_bs}).

We can observe that for the BS generation the novelty of the candidate CNs is lower than Top-$k$ (0.67-0.74 vs. 0.75-0.77) and the correlation with the novelty of the reference is weaker (0.53 vs. 0.59). This confirms the lower generalization ability with 
the deterministic decoding mechanism (as compared to the stochastic) that tends to produce generic and repetitive responses regardless of the semantic distances of the \textsc{loto} targets from the training data. 

\section{Automatic Post-Editing}\label{APE}

In the previous experiments we fine-tuned our models making resort to ${<}HS,CN{>}$ pairs alone. Still the \citet{bonaldi2021data} dataset contains additional information that can be useful for our task: i.\ e.\ the original GPT-2 generation before undergoing human post-editing.

Thus, as a final experiment, we propose to further improve the CN generation by moving from an end-to-end framework to a two stage pipeline, by decoupling CN generation from its `final refinement'.
In particular we propose the adoption of an Automatic Post-Editing (APE) stage in order to capture and utilize the nuances among the machine generated CNs and their human post-edited versions. APE, which is used for automatically correcting errors made by machine translation (MT) systems before performing actual human post-editing, has been an important tool for MT \citep{knight1994automated,do2021review}. 
Considering its effectiveness in MT, we hypothesize that building a pipeline with CN generation and APE could alleviate the requirement of the final manual post-editing \citep{allen2000toward,chatterjee2019findings} to achieve better constructed CNs.

To this end, we fine-tuned another instance of GPT-2 medium model specifically for the post-editing task using the ${<}HS,CN_{or},CN_{pe}{>}$ triplets\footnote{This is in line with the APE paradigm where the triplet is made of ${<}$\textit{source sentece}, $MT_{output}$, \textit{human post-edits}${>}$.}, where $CN_{or}$ and $CN_{pe}$ denote the CNs originally generated by an LM and their human post-edited versions, respectively. The triplets were then filtered by removing those for which $CN_{or} = CN_{pe}$. More details about the experiment settings can be found in Appendix \ref{sec:AppendixAPE}.

\begin{table}[htbp]
\begin{center}
\begin{tabular}{lrrr}
\hline
 Data &  $CN_{ape}$  & $CN_{or}$   &  N/A  \\
\hline
\citet{bonaldi2021data} &  26  &  14 & 60 \\
GPT-2 Top$\boldsymbol{_k}$ & 37 & 19 & 44 
\end{tabular}
\end{center} 
\caption{The human annotation results for the APE experiments in terms of average preference percentages.}
\label{tab:APE}
\end{table}

We have conducted two human evaluations over the subsets of: i) the $CN_{or}$ of the \citet{bonaldi2021data} test samples, ii) the CN outputs of the best model and decoding mechanism combination provided as the results of the first set of experiments, that yielded the top 50 Translation Error Rate (TER) \cite{snover2006study} scores with respect to the $CN_{or}$s. The two expert annotators were asked to state their preferences among the 2 randomly sorted CNs, $CN_{or}$ and $CN_{ape}$ (automatically post-edited output), for a given HS. The annotators were also allowed to decide on a tie. Results, shown in Table \ref{tab:APE}, indicate that, albeit there are often ties and only a subset of $CN_{or}$ is actually modified, when there is a preference, it is predominantly in favour of the automatically post-edited versions over the GPT-2 generated CNs (26\% vs. 14\% for the test set, and 37\% vs. 19\% for the GPT-2 Top$\boldsymbol{_k}$ generations, on average). Regarding the experiment results, we believe that APE is a highly promising direction to increase the efficacy of the CN generation models where generation quality and diversity is crucial, and considering that obtaining/enlarging expert datasets to train better models is not simple. 

\section{Conclusion}\label{Conclusion}

In this work, we focus on automatic CN generation as a downstream task. First, we present a comparative study to determine the performances and peculiarities of several pre-trained LMs and decoding mechanisms. We observe that the best results (in term of novelty and specificity) overall are achieved by the autoregressive models with stochastic decoding: GPT-2 with the Top$_k$ decoding mechanism, and DialoGPT with the combination Top$_{pk}$. At the same time deterministic decoding can be used when more generic yet `safer' CNs are preferred. 

Then, we investigate the performances of LMs in zero-shot generation for unseen targets of hate. Hence, we fine-tuned 5 different versions of GPT-2, leaving out the examples pertaining to one target at each turn. 
We find out that for each configuration/version, there is a subset of the training data which is more influential with respect to the generated data (i. e. a target that shares some commonalities with the test target that can be defined a-priori). 
Finally, we introduce an experiment by training an automatic post-editing module to further improve the CN generation quality. The notable human evaluation results paves the way for a promising future direction that decouples CN generation from its `final refinement'. 

\section*{Ethical Considerations}

Although tackling online hatred through CNs inherently protects the freedom of speech and has been proposed as a better alternative to the detect-remove-ban approaches, automatization of CN generation can still raise some ethical concerns and some measures must be taken to avoid undesired effects during research. Thus, we address the relevant ethical considerations and our remedies as follows:  

\paragraph{Annotation Guidelines:}
The well-being of the annotators was our top priority during the whole study. Therefore, we strictly followed the guidelines created for CN studies \citep{bonaldi2021data} that were adapted from \citep{vidgen2019challenges}. The human evaluations have been conducted with the help of two expert annotators in CNs. These experts were already trained for the CN generation task and employed for the work presented by \cite{bonaldi2021data}.

We further instructed them in the aims of each experiment, clearly explained the evaluation tasks, and then we exemplified proper evaluation of ${<}HS,CN{>}$ pairs using various types of CNs.

Most importantly, we limited the exposure to hateful content by providing a daily time limit of annotation. Concerning the demographics, due to the harmful content that can be found in the data, all annotators were adult volunteers, perfectly aware of the objective of the study.

\paragraph{Dataset.} We purposefully chose an expert-based dataset in order to avoid the risk of modeling the language of real individuals to (i) prevent any privacy issue, (ii) avoid to model  inappropriate CNs (e.g. containing abusive language) that could be produced by non-experts. The dataset also focuses on the CN diversity while keeping the HSs as stereotypical as possible so that our CN generation models have a very limited diversity on the hateful language, nearly precluding the misuse.

\paragraph{Computational Task.} CN generation models are not meant to be used in an autonomous way, since even the best models can still produce substandard CNs containing inappropriate or negative language. Instead, following a Human–computer cooperation paradigm, our focus is on building models that can be helpful to NGO operators by providing them diverse and novel CN candidates for their hate countering activities and speed up the manual CN writing to a certain extent. This approach also gives ground to some of the measures we used during evaluation (namely choose-or-not and is-best).  

\paragraph{Model Distribution.} In addition to the limited and simplified hateful content in the dataset we selected, we further reduce the risk of misuse by choosing a specific distribution strategy: i) we only make available the non-autoregressive models in order to eliminate the risk of using over-generation for hate speech creation, ii) we distribute such models only for research purposes and through a request based procedure in order to keep track of the possible users.

\bibliography{bibl}  
\bibliographystyle{acl_natbib}

\clearpage
\appendix

\section{Appendix}
\label{sec:appendix}
\setcounter{page}{1} 

\subsection{Fine-tuning details}\label{sec:AppendixA}

\noindent Table \ref{tab:training_details} summarizes the details of the training of each model employed in the first session of experiments. 

\begin{table}[htbp]
\begin{center}\resizebox{\columnwidth}{!}{
\begin{tabular}{@{}l|rrrrrrrrrr@{}}
\hline
\textbf{} & \multicolumn{1}{|c}{\textbf{BA}} & \multicolumn{1}{c}{\textbf{EP}} & \multicolumn{1}{c}{\textbf{PAR}} & \multicolumn{1}{c}{\textbf{LR}} & \multicolumn{1}{c}{\textbf{PER}} & \multicolumn{1}{c}{\textbf{TL}} & \multicolumn{1}{c}{\textbf{EL}} \\ \hline
BART (base)    & 4 & 4      & 139 M                                                                     & 2E-05                        & 24.659                                  & 2.358                                   & 2.417                                  \\
BERT Seq2Seq (base)     & 4 & 3      & 247 M                                                              & 3E-05                        & 11.209                                  & 2.845                                   & 3.205                                  \\
T5 (base)     & 2 & 3      & 223 M                                                                 & 5E-05                        & 10.9248                                 & 2.412                                   & 3.205                                  \\
DialoGPT (medium)    & 4 & 2     & {\ul \textbf{355 M}}                                                    & 5E-05        & {\ul \textbf{6.085}}                    & \textbf{1.425}                          & {\ul \textbf{1.806}}                   \\
GPT-2 (medium)    & 2 & 2     & {\ul \textbf{355 M}}                                                     & 5E-05        & \textbf{8.929}                          & {\ul \textbf{1.320}}                    & \textbf{2.189}                         
\end{tabular}}
\end{center}
\caption{The training details for all the models employed for the first collection of experiment: the batch size (BA), number of training epochs (EP), parameters (PAR), the learning rate (LR), perplexity (PER), training and evaluation loss (TL and EL).}
\label{tab:training_details}
\end{table}

\noindent Since LM sizes are very different for each model and since our main focus is not studying performances according to LM dimension growth, as a rule-of-thumb, we chose one version smaller than the large version of each model provided that they all have the same order of magnitude. This corresponds to the \textit{medium} versions for both DialoGPT and GPT-2, and \textit{base} versions for the other models.
GPT-2 and DialoGPT achieve the lowest perplexity, training and evaluation loss, thus indicating a slightly more successful fine-tuning, which are reflected in the evaluations throughout the study.

We conducted a hyper-parameter search during the training phase of each model using the search space: learning-rate:$\{1e-5, 2e-5, 3e-5, 4e-5, 5e-5\}$,
warm-up ratio:$\{0, 0.1\}$, batch-size:$\{2, 4\}$, epochs:$\{2, 3, 4, 5\}$. It has been conducted using Optuna, with 10 trials, optimized on minimizing the evaluation loss during training.

\subsection{Best models-decoding combination}\label{sec:AppendixD}

Here we discuss the results for the overlap and diversity metrics obtained on the Best$_{\text{\textsc{lm+d}}}$ generations (Table \ref{tab:ex1_best20_automatic-diversity}), and those calculated on the human evaluation subset (Tables \ref{tab:ex1_combo_tox_sd} and \ref{tab:ex1_combo_human}).

\paragraph{BART.}

BART performs well with the stochastic decoding methods, in particular: Top${_p}$ for overlap, diversity, syntactic metrics, and grammaticality; Top$_k$ for overlap metrics and toxicity, whereas Top$_{pk}$ is the best decoding approach on human evaluation and RR, and the second best on ROUGE and BLEU-1. On the contrary, BART does not achieve good results with deterministic approaches (i.\ e.\ BS). 

\begin{table}[htbp]
\begin{center}\resizebox{\columnwidth}{!}{\begin{tabular}{@{}l|rrrr|rr@{}}
\hline
\textbf{}   & \multicolumn{4}{|c}{\textbf{Overlap}}                                                                                                    & \multicolumn{2}{|c}{\textbf{Diversity}}                              \\ \hline
            & \multicolumn{1}{c}{\textbf{ROU}} & \multicolumn{1}{c}{\textbf{B-1}} & \multicolumn{1}{c}{\textbf{B-3}} & \multicolumn{1}{c}{\textbf{B-4}} & \multicolumn{1}{|c}{\textbf{RR}} & \multicolumn{1}{c}{\textbf{NOV}} \\
BART BS     & 0.2108                             & 0.2129                              & 0.0486                              & 0.0283                              & 21.1102                              & {\ul \textbf{0.5692}}                \\
BART Top$_{pk}$     & \textbf{0.2331}              &  \textbf{0.2300}               & 0.0605                              & 0.0365                              & {\ul \textbf{20.2645}}               & 0.5567                               \\
BART Top$_k$     & {\ul \textbf{0.2349}}                    & {\ul \textbf{0.2333}}                     & {\ul \textbf{0.0652}}               & {\ul \textbf{0.0385}}               & 20.6587                              & 0.5575                               \\
BART Top$_p$     & 0.2329                             & 0.2300                              & \textbf{0.0621}                     & \textbf{0.0374}                     & \textbf{20.5476}                     & \textbf{0.5586}                      \\ \hline
BERT BS     & 0.1735                             & 0.2108                              & 0.0249                              & 0.0113                              & 38.0349                              & 0.5864                               \\
BERT Top$_{pk}$     & \textbf{0.2034}                    & 0.2311                              & \textbf{0.0484}                     & \textbf{0.0231}                     & \textbf{23.4417}                     & 0.6098                               \\
BERT Top$_k$     & 0.2032                             & \textbf{0.2320}                     & 0.0483                              & 0.0229                              & {\ul \textbf{22.2546}}               & {\ul \textbf{0.6129}}                \\
BERT Top$_p$     & {\ul \textbf{0.2044}}              & {\ul \textbf{0.2366}}               & {\ul \textbf{0.0500}}               & {\ul \textbf{0.0244}}               & 23.6447                              & \textbf{0.6098}                      \\ \hline
T5 BS       & 0.2144                             & 0.2007                              & 0.0409                              & \textbf{0.0207}                     & 21.5518                              & 0.5827                               \\
T5 Top$_{pk}$       & {\ul \textbf{0.2236}}              & {\ul \textbf{0.2454}}               & {\ul \textbf{0.0466}}               & {\ul \textbf{0.0228}}               & 7.2996                               & 0.6715                               \\
T5 Top$_k$       & 0.2076                             & 0.2384                              & 0.0376                              & 0.0136                              & {\ul \textbf{5.3002}}                & {\ul \textbf{0.6922}}                \\
T5 Top$_p$       & \textbf{0.2159}                    & \textbf{0.2390}                     & \textbf{0.0430}                     & 0.0184                              & \textbf{6.8353}                      & \textbf{0.6743}                      \\ \hline
DialoGPT BS & {\ul \textbf{0.2192}}              & 0.2272                              & {\ul \textbf{0.0528}}               & {\ul \textbf{0.0312}}               & 21.6800                              & 0.5280                               \\
DialoGPT Top$_{pk}$ & \textbf{0.2132}                    & {\ul \textbf{0.2444}}               & \textbf{0.0437}                     & \textbf{0.0201}                     & 6.4158                               & 0.6737                               \\
DialoGPT Top$_k$ & 0.2023                             & 0.2302                              & 0.0320                              & 0.0134                              & {\ul \textbf{4.7278}}                & {\ul \textbf{0.6956}}                \\
DialoGPT Top$_p$ & 0.2093                             & \textbf{0.2397}                     & 0.0385                              & 0.0159                              & \textbf{6.1472}                      & \textbf{0.6740}                      \\ \hline
GPT-2 BS    & {\ul \textbf{0.2195}}              & 0.2132                              & {\ul \textbf{0.0516}}               & {\ul \textbf{0.0313}}               & 23.0605                              & 0.5402                               \\
GPT-2 Top$_{pk}$    & \textbf{0.2055}                    & {\ul \textbf{0.2342}}               & 0.0384                              & 0.0173                              & 6.5899                               & 0.6832                               \\
GPT-2 Top$_k$    & 0.1956                             & 0.2271                              & 0.0345                              & 0.0153                              & {\ul \textbf{4.7624}}                & {\ul \textbf{0.7022}}                \\
GPT-2 Top$_p$    & 0.2014                             & \textbf{0.2329}                     & \textbf{0.0388}                     & \textbf{0.0177}                     & \textbf{6.1944}                      & \textbf{0.6846}                      
\end{tabular}}
\end{center}
\caption{The results computed on the Best$_{\text{\textsc{m+d}}}$ generations (2500 CN for each model-decoding mechanism combination).}
\label{tab:ex1_best20_automatic-diversity}
\end{table}

\paragraph{BERT.}
With BS, BERT achieves the best or second best result on all human evaluation metrics, except for specificity. For BERT the best decoding is Top$_p$: it is the best performing on overlap metrics and the second best for novelty. It achieves good results both on syntactic metrics and human evaluation too. 

\begin{table}[htbp]
\begin{center}\resizebox{\columnwidth}{!}{
\begin{tabular}{l|r|rrr|r} \hline
            & \multicolumn{1}{|c}{\textbf{Toxicity}} & \multicolumn{3}{|c|}{\textbf{Syntactic metrics}}                                                         & \multicolumn{1}{l}{}           \\ 
             \hline
\textbf{}   & \multicolumn{1}{c}{\textbf{-}}        & \multicolumn{1}{|c}{\textbf{ASD}} & \multicolumn{1}{c}{\textbf{MSD}} & \multicolumn{1}{c}{\textbf{NST}} & \multicolumn{1}{|c}{\textbf{n}} \\
BART BS     & 0.4870                                       & 3.8919                           & 4.6757                           & {\ul \textbf{1.8919}}            & 37                             \\
BART Top$_{pk}$     & {\ul \textbf{0.3911}}                        & 4.3592                           & 4.9483                           & 1.6207                           & 58                             \\
BART Top$_k$     & \textbf{0.4021}                              & \textbf{4.3798}                  & \textbf{5.0656}                  & 1.7377                           & 61                             \\
BART Top$_p$     & 0.4263                                       & {\ul \textbf{4.5038}}            & {\ul \textbf{5.0909}}            & \textbf{1.7727}                  & 44                             \\ \hline
BERT BS     & {\ul \textbf{0.3954}}                        & 4.5556                           & 5.3750                           & 1.9167                           & 24                             \\
BERT Top$_{pk}$     & \textbf{0.4026}                              & {\ul \textbf{5.2299}}            & 6.2069                           & 2.1379                           & 58                             \\
BERT Top$_k$     & 0.4157                                       & 4.8969                           & {\ul \textbf{6.2969}}            & {\ul \textbf{2.5625}}            & 64                             \\
BERT Top$_p$     & 0.4032                                       & \textbf{5.1019}                  & \textbf{6.2963}                  & \textbf{2.2593}                  & 54                             \\ \hline
T5 BS       & 0.4127                                       & 4.4844                           & 4.6562                           & 1.3438                           & 32                             \\
T5 Top$_{pk}$       & {\ul \textbf{0.3211}}                        & {\ul \textbf{4.7754}}            & 5.3768                           & \textbf{1.7826}                  & 69                             \\
T5 Top$_k$       & \textbf{0.3441}                              & 4.6767                           & \textbf{5.4200}                  & 1.7400                           & 50                             \\
T5 Top$_p$       & 0.3934                                       & \textbf{4.7245}                  & {\ul \textbf{5.5918}}            & {\ul \textbf{1.8367}}            & 49                             \\ \hline
DialoGPT BS & 0.3635                                       & 4.2340                           & 5.1277                           & 1.8723                           & 47                             \\
DialoGPT Top$_{pk}$ & \textbf{0.3361}                              & 4.7264                           & 5.5094                           & 1.7547                           & 53                             \\
DialoGPT Top$_k$ & 0.3482                                       & {\ul \textbf{4.9333}}            & {\ul \textbf{6.1778}}            & {\ul \textbf{2.0000}}            & 45                             \\
DialoGPT Top$_p$ & {\ul \textbf{0.3274}}                        & \textbf{4.7970}                  & \textbf{5.5273}                  & \textbf{1.9636}                  & 55                             \\ \hline
GPT-2 BS    & 0.3540                                       & {\ul \textbf{4.8901}}            & 5.3617                           & 1.4468                           & 47                             \\
GPT-2 Top$_{pk}$    & {\ul \textbf{0.3119}}                        & 4.2530                           & 5.4182                           & 2.4000                           & 55                             \\
GPT-2 Top$_k$    & \textbf{0.3416}                              & \textbf{4.6771}                  & {\ul \textbf{5.8627}}            & {\ul \textbf{2.5686}}            & 51                             \\
GPT-2 Top$_p$    & 0.3659                                       & 4.5663                           & \textbf{5.7447}                  & \textbf{2.4894}                  & 47                             
\end{tabular}}
\end{center}
\caption{The results of the toxicity and the syntactic metrics calculated on the subset employed for the human evaluation and grouped by each combination of model and decoding mechanism. The size of each group is showed in the column ``n''.}
\label{tab:ex1_combo_tox_sd}
\end{table}

\paragraph{T5.}
For T5, Top$_{pk}$ is the best decoding mechanism. It records the best results for overlap metrics and toxicity, and it has good results on syntactic and human evaluation metrics. For what regards Top$_k$, it is the best for diversity, while Top$_p$ is good on the syntactic metrics. BS achieves good results on human evaluation, except for specificity and is-best. 

\paragraph{GPT-2.} 
With Top$_{pk}$, GPT-2 performs well on ROUGE, BLEU-1, suitableness, grammaticality, and choose-or-not. With Top$_p$, GPT-2 records the second best result on BLEU scores and diversity metrics. With BS the model has the best performance on overlap metrics (except BLEU-1), and on suitableness, grammaticality, and choose-or-not, but it has also the worst results on diversity metrics. Above all, Top$_k$ is the decoding achieving the best compromise, reaching the best results for the diversity metrics, and with a superior specificity score (3.15) that is corroborated by the good performance on the other human evaluation metrics.

\paragraph{DialoGPT.} 
Top$_k$ performs best with diversity metrics and specificity; it records the second highest score on grammaticality. Top$_p$ has the second best result on diversity metrics and BLEU scores. BS is the best on overlap metrics (except BLEU-1), and also on almost all human evaluation metrics: it is the worst on specificity and on diversity metrics. 

Top$_{pk}$ is the one working best with DialoGPT, since it reaches very good scores with human and overlap metrics, and this does not invalidate diversity, for which it ranks 3rd out of 4.

\begin{table}[htbp]
\begin{center}\resizebox{\columnwidth}{!}{
\begin{tabular}{@{}l|rrrrr|r@{}}
\hline
\multicolumn{1}{r}{\textbf{}} & \multicolumn{5}{|c}{\textbf{Human evaluation}}    & \multicolumn{1}{|c}{} \\ \hline
\multicolumn{1}{r}{\textbf{}} & \multicolumn{1}{|c}{\textbf{SUI}} & \multicolumn{1}{c}{\textbf{SPE}} & \multicolumn{1}{c}{\textbf{GRM}} & \multicolumn{1}{c}{\textbf{CHO}} & \multicolumn{1}{c}{\textbf{BEST}} & \multicolumn{1}{|c}{\textbf{n}} \\
BART BS   & 3.7568                  & 2.5270                  & 4.9459                 & 0.8108                     & 0.2297                & 37             \\
BART Top$_{pk}$   & \textbf{3.7931}         & {\ul \textbf{2.6121}}   & \textbf{4.9483}        & \textbf{0.8534}            & {\ul \textbf{0.3707}} & 58             \\
BART Top$_k$   & {\ul \textbf{3.9672}}   & \textbf{2.5410}         & 4.9016                 & {\ul \textbf{0.8607}}      & \textbf{0.2951}       & 61             \\
BART Top$_p$   & 3.5682                  & 2.5114                  & {\ul \textbf{4.9659}}  & 0.8182                     & 0.1477                & 44             \\ \hline
BERT BS   & {\ul \textbf{3.5208}}   & 2.5208                  & {\ul \textbf{4.7917}}  & {\ul \textbf{0.7708}}      & \textbf{0.1250}       & 24             \\
BERT Top$_{pk}$   & \textbf{3.1810}         & 2.5776                  & \textbf{4.2328}        & 0.7155                     & 0.1121                & 58             \\
BERT Top$_k$   & 3.0312                  & \textbf{2.7031}         & 4.1562                 & 0.6797                     & 0.1016                & 64             \\
BERT Top$_p$   & 3.0370                  & {\ul \textbf{2.7130}}   & 4.1296                 & \textbf{0.7407}            & {\ul \textbf{0.1574}} & 54             \\ \hline
T5 BS     & {\ul \textbf{3.5781}}   & 2.2812                  & {\ul \textbf{4.8438}}  & {\ul \textbf{0.7656}}      & 0.0781                & 32             \\
T5 Top$_{pk}$     & \textbf{2.8841}         & {\ul \textbf{2.4928}}   & 4.5870                 & \textbf{0.6667}            & \textbf{0.1014}       & 69             \\
T5 Top$_k$     & 2.4600                  & 2.3200                  & 4.6400                 & 0.5600                     & 0.0500                & 50             \\
T5 Top$_p$     & 2.8163                  & \textbf{2.4388}         & \textbf{4.7449}        & 0.6122                     & {\ul \textbf{0.1224}} & 49             \\\hline
DialoGPT BS                   & {\ul \textbf{4.1596}}             & 2.6064                            & {\ul \textbf{4.9894}}              & {\ul \textbf{0.8511}}               & {\ul \textbf{0.3085}}             & 47                             \\
DialoGPT Top$_{pk}$                   & \textbf{3.3679}                   & \textbf{2.8019}                   & 4.8396                             & \textbf{0.7830}                     & \textbf{0.2736}                   & 53                             \\
DialoGPT Top$_k$                   & 3.1333                            & {\ul \textbf{2.9222}}             & \textbf{4.8556}                    & 0.7333                              & 0.2111                            & 45                             \\
DialoGPT Top$_p$                   & 2.9727                            & 2.7000                            & 4.8455                             & 0.7091                              & 0.1909                            & 55                             \\ \hline
GPT-2 BS                      & {\ul \textbf{4.3085}}             & 2.5000                            & {\ul \textbf{4.9681}}              & {\ul \textbf{0.8830}}               & {\ul \textbf{0.2766}}             & 47                             \\
GPT-2 Top$_{pk}$                      & \textbf{3.4909}                   & 2.8000                            & \textbf{4.8727}                    & \textbf{0.8273}                     & \textbf{0.2273}                   & 55                             \\
GPT-2 Top$_k$                      & 3.0392                            & {\ul \textbf{3.1471}}             & 4.8431                             & 0.7255                              & 0.2549                            & 51                             \\
GPT-2 Top$_p$                      & 3.4362                            & \textbf{3.0638}                   & 4.7872                             & 0.7447                              & 0.3298                            & 47                            
\end{tabular}}
\end{center}
\caption{For each model-decoding mechanism combination, these are the results of the metrics for the human evaluations. The size of each combination is showed in the column ``n".}
\label{tab:ex1_combo_human}
\end{table}

\subsection{Examples of the generations}\label{sec:AppendixC}

\begin{quote}
    \textbf{Hate speech}: Islam is a violent religion that promotes terror and war.   
\end{quote}

\begin{quote}
    \textbf{DialoGPT - BS}: \textit{The vast majority of Muslims in the world are peaceful and friendly people. I don't see how you can say such a thing about them.}
\end{quote}

\begin{quote}
    \textbf{BART - Top$\boldsymbol{_k}$}: \textit{The vast majority of Muslims are peaceful people who have lived in peace and harmony for many years. How can you say this about a religion that promotes terror and war?}
\end{quote}

\begin{quote}
    \textbf{BERT - Top$\boldsymbol{_k}$}: \textit{islam is a religion of peace, and it is not a peaceful religion of violence. it is about peace and understanding of the world's problems, and peace is about bringing peace and peace.}
\end{quote}

\begin{quote}
    \textbf{T5 - Top$\boldsymbol{_p}$}: \textit{The vast majority of Muslims in the UK are peaceful people who share many values and respect.}
\end{quote}

\begin{quote}
    \textbf{GPT-2 - Top$\boldsymbol{_p}$}: \textit{This is not true, the vast majority of Muslims is peaceful and peaceful Muslims are happy to live in any country they wish. They do not need to be a part of a hostile state and they should be treated as equals in society.}
\end{quote}

\noindent In these examples we can observe that the CN obtained with GPT-2, albeit not perfect, is the most complete and accurate. Those generated with T5 and DialoGPT are suitable as well. On the other hand, the CNs obtained with BART and BERT present some problems: the former has a clear contradiction in it and the latter is illustrative of some of the problems detected in BERT's generations such that it is syntactically complex but very vague and repetitive, and it ends up not really conveying a clear message. 

\begin{quote}
    \textbf{Hate speech}: The migrant crisis is not the fault of Western Europeans. Migrants are to blame for immigration.
\end{quote}

\begin{quote}
    \textbf{DialoGPT - Top$\boldsymbol{_k}$}: \textit{What if an entire religion was to blame only for the migration crisis? So you think a large number of Muslims are responsible for all the problems of our world?}
\end{quote}

\begin{quote}
    \textbf{BART - Top$\boldsymbol{_p}$}: \textit{The migrant crisis is caused by a lack of integration and lack of knowledge of the world.}
\end{quote}

\begin{quote}
    \textbf{BERT - BS}: \textit{many migrants come here to work and provide for their families. they are our brothers and sisters, and we should welcome them.}
\end{quote}

\begin{quote}
    \textbf{T5 - Top$\boldsymbol{_k}$}: \textit{What evidence do you have for the evidence that migrants are to blame? Not all migrants have done anything wrong.}
\end{quote}

\begin{quote}
    \textbf{GPT-2 - Top$\boldsymbol{_k}$}: \textit{The EU has put in place laws to combat illegal immigration and to prevent crimes committed by migrants. They have a duty to help stop crime and do what is necessary to solve the problems within their borders. Migrants are very intelligent and do what they can to support their families while they study and work.}
\end{quote}

\noindent Once again, GPT-2 and DialoGPT generations where chosen as the best CNs by the annotators. Moreover, these CNs are all generated via Top$_k$ decoding mechanism. BART and T5 generated CNs are acceptable while requiring some post-editing to be employed. Finally, BERT's output is still very vague: we can notice the presence of the recurring expression \textit{our brothers and sisters}.

\subsection{Additional material for \textsc{Loto} experiments}\label{sec:AppendixB}

Table \ref{tab:LOTO_coverage} displays the distribution of the examples with respect to the targets, in the reference dataset and in the configurations for the \textsc{loto} experiments (Section \ref{subsec:loto-experiment}).

\begin{table}[htbp]
\begin{center}
\resizebox{\columnwidth}{!}{\begin{tabular}{@{}l|rr@{}}
\hline
Target   & Samples in original  & Samples in \textsc{loto} \\ 
& dataset & experiment \\
\hline
\texttt{JEWS}     & 594                        & 600          \\
\texttt{LGBT+}    & 617                        & 600             \\
\texttt{MIGRANTS} & 957                        & 600             \\
\texttt{MUSLIMS}  & 1335                       & 600             \\
\texttt{WOMEN}    & 662                        & 600             \\
\texttt{DISABLED} & 220                        & 220             \\
\texttt{POC}      & 352                        & 352             \\
\texttt{other}    & 266                        & 157             \\  \hline
Total             & 5003                       & 3729            
\end{tabular}}
\end{center}
\caption{The targets coverage in the reference dataset \citep{bonaldi2021data} and in the \textsc{loto} configurations.}
\label{tab:LOTO_coverage}
\end{table}

\begin{table}[htbp]
\begin{center}\resizebox{\columnwidth}{!}{
\begin{tabular}{l|rrrrr}
\hline
generation &   \texttt{JEWS}&  \texttt{LGBT+}&  \texttt{MIGRANTS}&  \texttt{MUSLIMS}&  \texttt{WOMEN} \\
training &        &        &           &          &        \\
\hline
\texttt{JEWS}     &  - &  0.775 &     0.780 &    \textbf{0.761} &  0.780 \\
\texttt{LGBT+}    &  0.781 &  - &     0.783 &    0.765 &  \textbf{0.763} \\
\texttt{MIGRANTS}&  0.782 &  0.775 &     - &    0.764 &  0.777 \\
\texttt{MUSLIMS}  &  0.775 &  0.770 &     0.769 &    - &  0.776 \\
\texttt{WOMEN}    &  0.789 &  0.771 &     0.783 &    0.775 &  - 
\end{tabular}}
\end{center}
\caption{The novelty of the reference CNs in the data from \citet{bonaldi2021data} (\textit{generation}) with respect to the training data for the \textsc{loto} models (\textit{training}).}
\label{tab:LOTO_MTCONAN_Novelty}
\end{table}

Table \ref{tab:LOTO_MTCONAN_Novelty} presents the detailed results for the novelty of the reference CNs discussed in Section \ref{subsec:loto-results}, while the RR for the CNs generated with the \textsc{loto} models and for the reference CNs are shown in Table \ref{tab:LOTO_RR}. The rankings for these two RR computations are the same, and the ranges are almost overlapping. This means that leaving one target out does not impact the intra-corpora repetitiveness: instead, the CNs generated with a \textsc{loto} model gain a lower RR than the reference CNs. 

For the target \texttt{MUSLIMS} a high RR is recorded, both in candidate and in the reference CNs. A high repetitiveness in the data for this target can contribute to the good results observed on overlap metrics too (Table \ref{tab:LOTO_results} in Section \ref{subsec:loto-results}): it is easier that two outputs are similar if they use a limited and repeated number of words.

\begin{table}[htbp]
\begin{center}
\resizebox{\columnwidth}{!}{\begin{tabular}{@{}l|rr@{}}
\hline
{}Target &  RR \textit{reference CN}  & RR \textit{candidate CN}  \\
\hline
 \texttt{JEWS}&            5.071 &                4.796 \\
\texttt{LGBT+}&            4.489 &                 4.620   \\
\texttt{MIGRANTS}&            4.381  &             4.707\\
\texttt{MUSLIMS} &            5.244  &              5.314 \\
\texttt{WOMEN} &            4.547  &                4.632

\end{tabular}}
\end{center}
\caption{The RR computed on the reference CN (pertaining the test set) and on the CN generated with the \textsc{loto} models.}
\label{tab:LOTO_RR}
\end{table}

\subsection{APE Experiment Details}\label{sec:AppendixAPE}
The dataset by \citep{bonaldi2021data} contains three versions of the same CN: the original CN generated by a GPT-2 model (CN$_{or}$), the expert post-edited versions obtained during the human-in-the-loop cycles (CN$_{pe*}$), and the final version rechecked by NGO experts (CN$_{pe}$). 

For fine-tuning our APE model, we have thus used the triplets ${<}HS,CN_{or},CN_{pe}{>}$ and ${<}HS,CN_{pe*},CN_{pe}{>}$. In this way, we managed to roughly double the number of the post-edit training samples, which is highly beneficial for a better model. 
When we filtered the triplets with a positive TER score between CN$_{ed}$ and CN$_{pe}$, or CN$_{or}$ and CN$_{pe}$, we obtained 4185 training, 596 test, and 568 validation samples following the partition used in the first set of experiments as described in Section \ref{sec:dataset}.
Finally, the best fine-tuning configuration of the GPT-2 medium model for APE was obtained with a learning rate of 2e-5 for 3 epochs resulting in 3.34 train loss and 1.23 eval loss. 

\end{document}